\documentclass[11pt]{article}

\usepackage[T1]{fontenc}
\usepackage[utf8]{inputenc}
\usepackage{lmodern}
\usepackage{amsmath,amssymb}
\usepackage{graphicx}
\usepackage{authblk}
\usepackage{algorithm}
\usepackage{algpseudocode}
\usepackage{booktabs}
\usepackage{subcaption}
\usepackage{multirow}
\usepackage{tabularx}
\usepackage{makecell}
\usepackage{xcolor}
\usepackage{placeins}
\usepackage{float}
\usepackage[title]{appendix}
\usepackage[numbers]{natbib}
\usepackage[colorlinks=true,linkcolor=blue,citecolor=blue,urlcolor=blue]{hyperref}
\usepackage[margin=1in]{geometry}

\graphicspath{{./}{figs/}}
\title{Stable but Wrong: When Learning Stabilizes Away from the Truth}
\author[1,2]{Zhipeng Zhang}
\author[4]{Fujie Ren}
\author[1]{Kai Li}
\author[3]{Bo Qiu}
\affil[1]{China Mobile Research Institute, Beijing, China}
\affil[2]{China Mobile GBA (Greater Bay Area) Innovation Institute, Guangzhou, China}
\affil[3]{University of Science and Technology Beijing, Beijing, China}
\affil[4]{University of Electronic Science and Technology of China, Chengdu, China}
\date{}

\hypersetup{
  pdftitle={Stable but Wrong: When Learning Stabilizes Away from the Truth},
  pdfauthor={Zhipeng Zhang; Fujie Ren; Kai Li; Bo Qiu}
}

\begin{document}
\maketitle
\vspace*{-2.2em}
\begin{center}
{\small Contact: Zhipeng Zhang, \texttt{zzp\_zzp2002@aliyun.com}}
\end{center}
\vspace{-0.8em}

\begin{abstract}
Stable training is often treated as evidence that learning is succeeding, but stability characterizes optimization behavior rather than correctness relative to an external objective. We study what happens when the signal being optimized remains persistently biased. We define \textbf{Stable but Wrong (SBW)} as a learning state in which optimization remains active or stabilizes while the learned outcome stays systematically displaced from an independently defined objective. A minimal theoretical model shows that persistent systematic bias can separate a stable convergence point from the true objective even under standard conditions for stable convergence. We then introduce static and dynamic bias in reinforcement learning, supervised learning, and continual fine-tuning of large language models. Across these settings, optimization of the current learning signal continues or stabilizes while correctness relative to the external objective declines or remains displaced. Finally, two controlled challenge-correction experiments with external ground truth test whether a prespecified error can be exposed and changed. In controlled scientific discovery and CIFAR-10, targeted evidence corrects or reduces established errors more effectively than equal-budget general evidence, with no evident harm to paired correct states under the tested protocols. These experiments do not provide a general method for automatically discovering unknown errors. Overall, SBW identifies a basic limit of convergence and other internal optimization metrics as evidence of correctness in systems that learn through persistent bias and feedback.
\end{abstract}

\section{Introduction}

Falling training loss, convergence, and stable performance are often taken as signs of successful learning. In many standard settings, they accompany improved task performance. Yet they do not establish that a model has learned the true objective.

\begin{figure}[H]
\centering
\includegraphics[width=\textwidth]{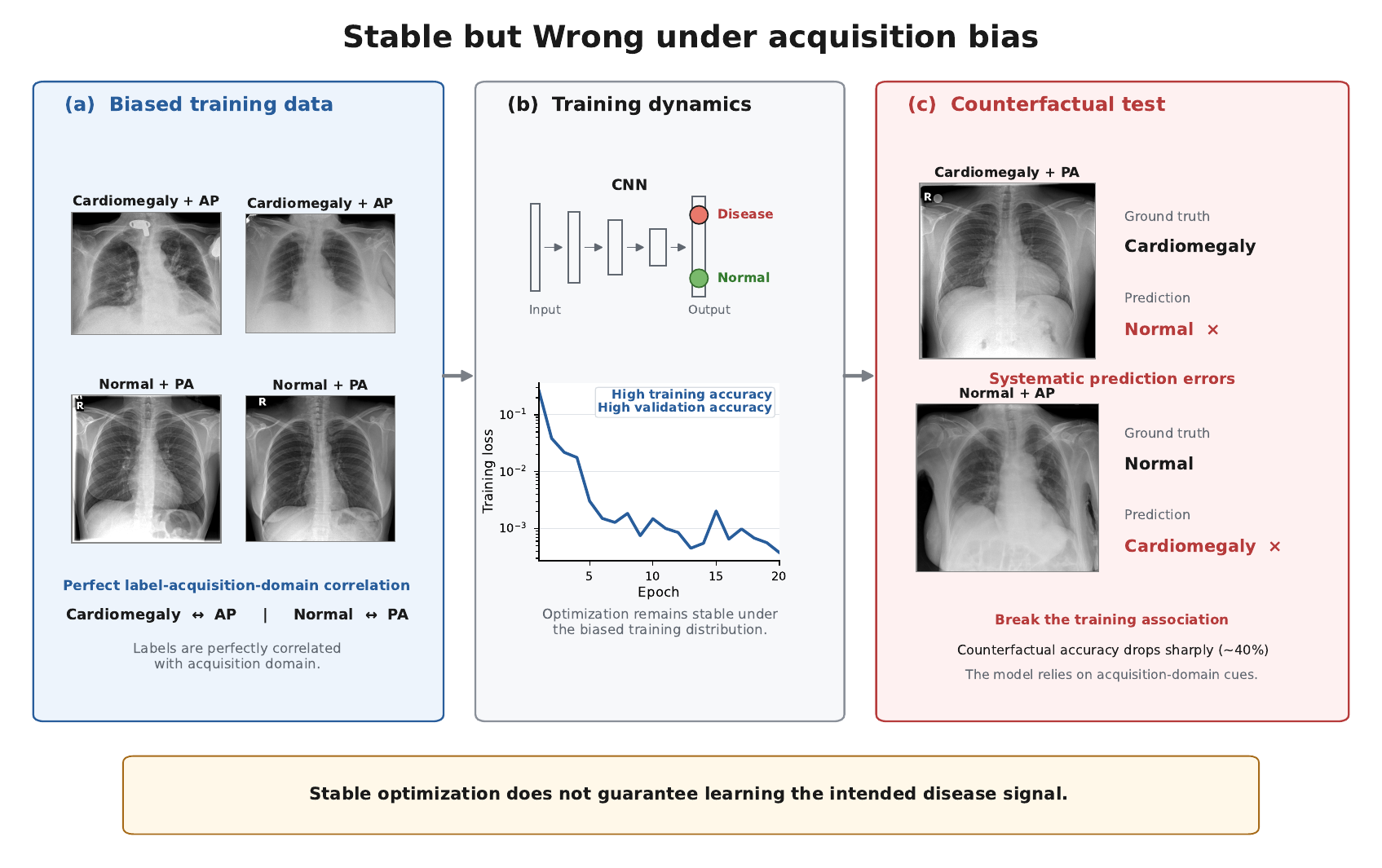}
\caption{
\textbf{Stable but Wrong in supervised learning.}
A controlled PadChest example. Controlled sample selection creates perfect label--acquisition-domain collinearity between Cardiomegaly--AP and Normal--PA. The images and AP/PA metadata come from real clinical data; the perfect collinearity is experimentally constructed and does not represent the original PadChest population. From left to right: training data, training dynamics, and a counterfactual test that breaks the association. The disease labels are retained while the acquisition domains are exchanged.
}
\label{fig:intro_medical_sbw}
\end{figure}

Medical imaging offers an intuitive example. Prior work has found that artificial-intelligence models can achieve high accuracy in disease-recognition tasks while relying mainly on features unrelated to the disease itself, such as hospital source, imaging equipment, or acquisition protocol \cite{RadiologyShortcut2024,DeGrave2021,Lapuschkin2019}. Figure~\ref{fig:intro_medical_sbw} presents a controlled PadChest example \cite{Bustos2020PadChest}. Using real clinical images and their original disease labels, we select samples to create a systematic association between disease class and X-ray acquisition domain. The model trains with a normally decreasing loss and reaches 97.9\% accuracy on a validation set that preserves this association. When the test data break the association between disease and acquisition domain, however, accuracy falls to about 40\%. The model appears to recognize disease, but may instead rely primarily on how the X-ray was acquired. We do not alter image content or supply incorrect disease labels; the bias arises from the association between labels and acquisition conditions. This perfect association is constructed through sample selection and does not represent the original PadChest population.

Hidden errors of this kind are not limited to medicine. Consider an AI system designed for early fire detection. If most fire photographs in its training data were taken after fire engines arrived, while fire engines rarely appear in ordinary scenes, the model might treat a fire engine as an important cue even when every photograph has a correct label. It could improve its training accuracy while missing a fire that has just started, before firefighters arrive. This hypothetical scenario shares a mechanism with acquisition bias in medical imaging: the model learns a real statistical association in the data and performs well on it, but that association may not reliably support the intended task.

These examples expose a basic distinction: correct observations and labels do not guarantee that the learned pattern reflects the task objective. Even without obvious data errors, training signals may encode proxy associations that ordinary learning does not detect. As learning continues, the model may adapt more fully to those associations, allowing an erroneous basis for judgment to become a stable outcome.

The practical risk is that hospital source, equipment, imaging protocols, and other data-collection or environmental factors can become stable but unreliable proxies for the task without anyone assigning incorrect labels. If these associations persist in training and validation data, a model may retain favorable internal metrics until the association is broken, the environment changes, or an independent evaluation is introduced. Reliability therefore depends on whether learned patterns hold beyond the conditions that created the original association.

The issue takes on added importance in autonomous scientific-discovery systems. In recent years, autonomous materials-discovery systems such as A-Lab have combined machine-learning predictions, candidate selection, and automated experimental platforms to participate in materials discovery \cite{Szymanski2023}. Unlike conventional learning from a fixed dataset, such systems use experiments to determine which observations become available next. As AI takes a larger role in forming scientific knowledge, an important question follows: when a scientific model's predictions become stable, does that mean it has discovered the correct scientific regularity?

\begin{figure}[H]
\centering
\includegraphics[width=0.8\textwidth]{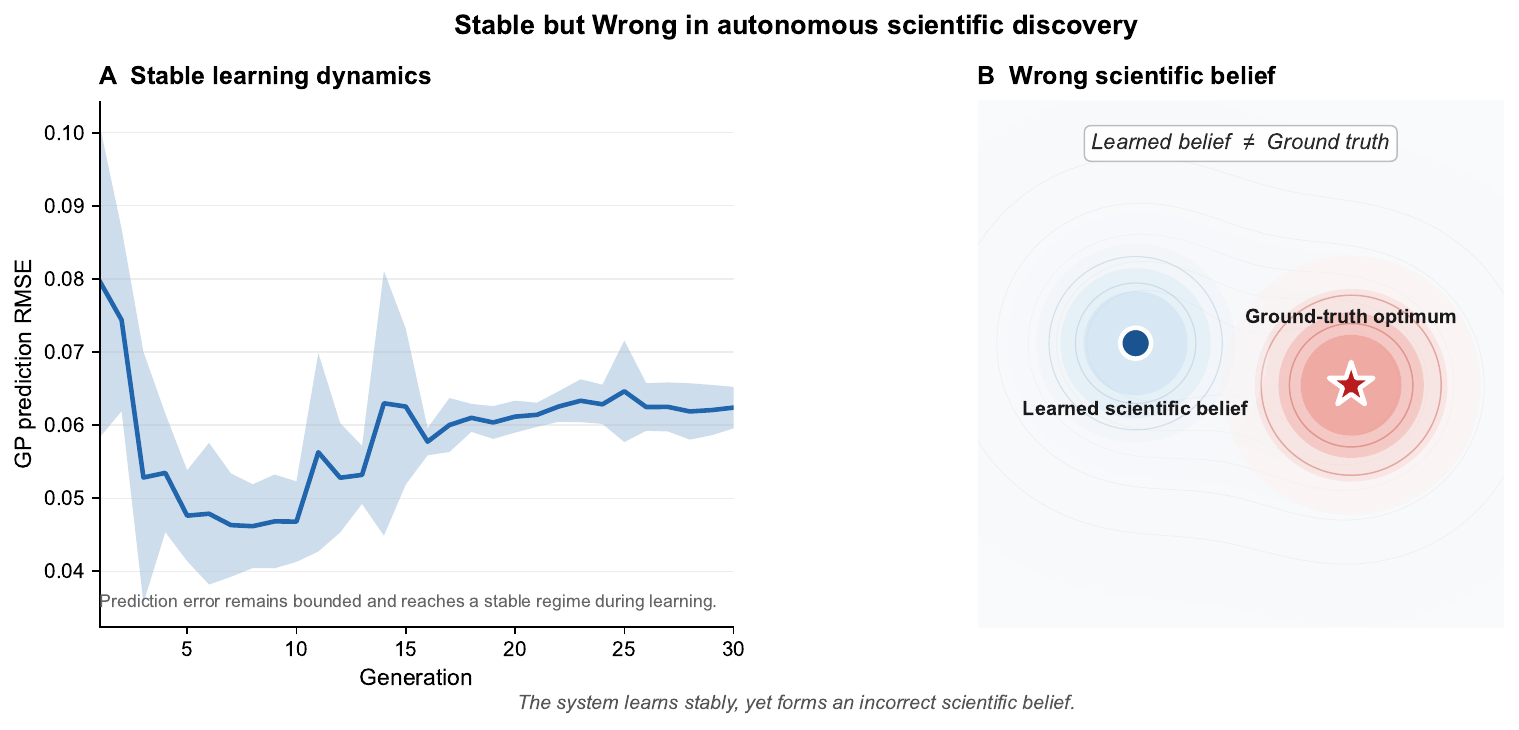}
\caption{
\textbf{Stable but Wrong in synthetic autonomous discovery.}
The Synthetic Discovery System constructed in this study uses neither real A-Lab data nor an actual experimental platform. The figure compares the learned belief of a Bayesian-optimization surrogate model with the known oracle landscape: after the belief stabilizes, the region identified as optimal remains displaced from the true global optimum.
}
\label{fig:intro_autolab_sbw}
\end{figure}

To examine this question when the true objective is known, we construct a \textbf{Controlled Synthetic Autonomous Discovery System} (hereafter, the \emph{Synthetic Discovery System}). The system runs Bayesian optimization on a five-dimensional synthetic objective with a known oracle landscape, updates its surrogate model from a limited set of observations, and selects subsequent experiments using its current predictions. Figure~\ref{fig:intro_autolab_sbw} shows that, across five random seeds, prediction error on held-out candidate points remains bounded and enters a relatively stable regime, while the optimal region identified by the learned scientific belief remains systematically displaced from the true global optimum. The stability of prediction error describes predictions on held-out candidates; it does not mean that the system has found the global optimum. This experiment uses a synthetic system with a known ground-truth objective, not a failure case from the real A-Lab. It shows that a learning system may model its observed data stably without forming a correct scientific understanding. In autonomous discovery, the potential consequence of this discrepancy may extend beyond an incorrect prediction: the system may continue to maintain an incorrect scientific judgment.

Bias need not be artificially injected or appear as an optimization anomaly. Data collection and measurement can introduce observation bias; labels, rewards, and proxy objectives can introduce supervision or objective bias; and selection, feedback, or reuse of model outputs can introduce selection, feedback, or recursive bias \cite{Hardt2025Performative,Mehrabi2022BiasSurvey,Dogra2024ShortcutSurvey}. These biases may be static or may evolve with the learner's state, actions, and updates. Their mechanisms differ, and not every bias produces error. If a displaced signal with stable statistical structure continues to shape learning, the system may become better at optimizing that signal without moving closer to the true objective.

We call a state in which learning proceeds normally or becomes stable while its outcome remains systematically displaced from an external objective \textbf{Stable but Wrong (SBW)}. SBW describes a learning state that different persistent-bias mechanisms can produce, rather than a new source of bias. It separates learning stability from learning correctness: convergence can show adaptation to the current signal, but cannot establish that the signal faithfully represents the external objective. Under some conditions, ordinary learning can gradually form and retain an error.

SBW is related to, but distinct from, established learning failures. Shortcut learning and Clever Hans behavior describe models that rely on proxy cues unrelated to the task \cite{Geirhos2020Shortcut,Brown2023Shortcut,Kauffmann2025CleverHans,Lapuschkin2019}; confirmation bias in self-training can reinforce prediction errors; and recursive training can degrade data distributions through repeated reuse of model-generated data \cite{nature2024collapse}. These studies address different causes and consequences. We ask whether persistent bias across learning updates can yield a shared state in which the process stabilizes while its outcome remains displaced from the true objective.

The question matters as AI systems increasingly update through feedback, actions, and self-generated data. Their future learning signals may depend on their own earlier outputs, allowing bias to persist or evolve without an obvious training anomaly. Stability is therefore an unreliable proxy for correctness in long-running learning systems.

We ask how persistent bias interacts with ordinary learning dynamics to make error a stable outcome. A minimal theoretical model shows that persistent systematic bias can separate a stable convergence point from the true objective even under standard conditions for stable convergence (proof in Appendix~\ref{app:theory}). This establishes the existence of SBW in a simple setting and distinguishes stable optimization from correctness relative to the true objective.

We then test for this state under static and dynamic bias in reinforcement learning, supervised learning, and continual fine-tuning of large language models. Although the systems differ in data, architecture, and optimization, each shows a separation between optimization and external correctness: reinforcement-learning policy updates remain active without evident numerical divergence as true returns deteriorate; supervised-learning loss continues to fall or, after a brief increase when bias is introduced, declines; and language-model training loss stabilizes as performance on the original task declines. Active, non-divergent policy updates do not independently establish strict convergence.

Finally, two controlled challenge--correction experiments—in synthetic autonomous discovery and CIFAR-10 Structural Patch classification—test whether external evidence can confirm and correct or reduce an established error. With external ground truth and a prespecified judgment to test, targeted evidence is more effective than equal-budget general data additions. We observe no evident harm to paired correct states under the tested protocol. This is a bounded validation, not a general method for automatically discovering unknown errors.

The theory and experiments characterize SBW as a state in which persistent bias allows learning to stabilize away from an external objective. A system can optimize its current signal successfully while forming incorrect knowledge or decisions. Assessing its reliability therefore requires independent evidence that its conclusions match the objective.

\section{Results}


\subsection{Persistent bias in learning signals can produce Stable but Wrong states}

We test whether Stable but Wrong (SBW) states recur across reinforcement learning (RL), supervised learning (SL), and continual fine-tuning of large language models (LLMs). The experiments examine static bias and bias that evolves with the learner's state.

\begin{figure}[H]
\centering
\includegraphics[width=0.77\linewidth]{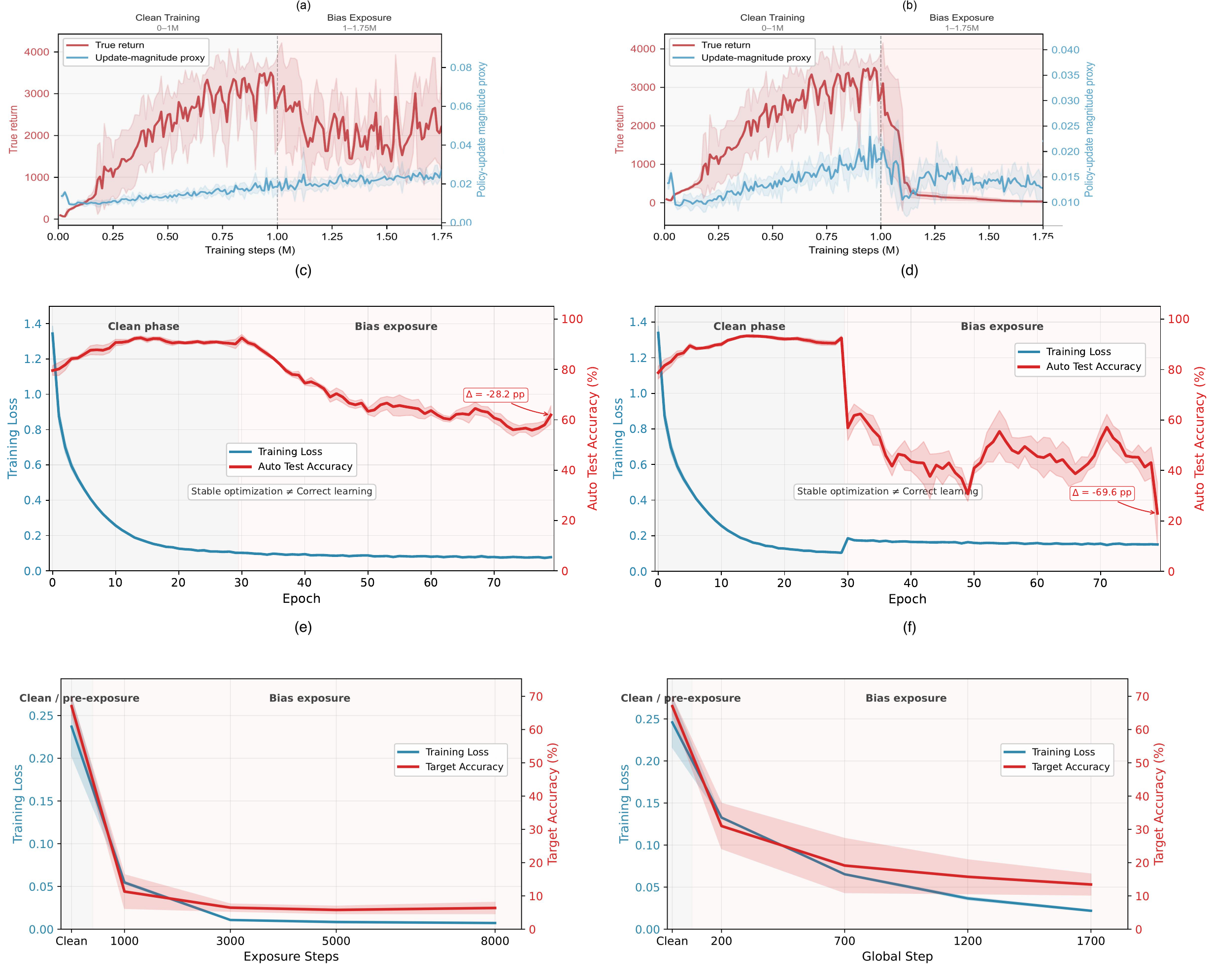}
\caption{
Learning processes and correctness across paradigms under persistent bias. (a,b) Static and dynamic bias in PPO on Hopper-v4: training includes Clean Learning and Bias Exposure and ends at the end of Bias Exposure (1.75M steps); red, true return; blue, a proxy for policy-update magnitude; shading, standard deviation across five seeds. This proxy indicates only that policy updates remain active without explicit divergence; it does not prove strict convergence. (c,d) Structural and prediction-dependent label bias in CIFAR-10/ResNet-18 (epochs 30--79): blue, training loss; red, clean target-class accuracy. Accuracy curves are smoothed over five epochs within each stage, while stage endpoints show raw means. Blue shading denotes standard deviation across five seeds; red shading denotes standard error across five seeds. Changes shown in the figure use the difference between raw means at the final Clean epoch and final Bias epoch; the Clean reference in the appendix instead uses the mean over the final five epochs. (e,f) Static and dynamic bias in Qwen2.5-7B-Instruct on GSM8K: blue, windowed training loss; red, final-answer accuracy on the Target subset; shading, standard deviation across five seeds.
}
\label{fig:stable_but_wrong}
\end{figure}

Static bias persists during learning but does not depend on the model's state. Examples include acquisition-related shortcuts in medical imaging \cite{Geirhos2020Shortcut} and fixed erroneous supervision during continual fine-tuning \cite{Dogra2024ShortcutSurvey}. Dynamic bias depends on the learner's state and can therefore propagate or evolve through subsequent updates, as when model predictions shape later training data or feedback \cite{Hardt2025Performative}.

Controlled bias injections let us specify the bias and retain an independent ground truth or oracle for evaluating correctness. In real systems, bias can arise through data collection, feedback, or model interaction, while its source, magnitude, and the true objective may not all be observable. The controlled designs therefore test whether a system can continue learning while its signal remains displaced from the objective, with other learning conditions held constant.

The three settings capture distinct update mechanisms: fixed datasets in supervised learning \cite{Krizhevsky2009CIFAR,He2016ResNet}, environment-driven policy updates in reinforcement learning \cite{Schulman2017PPO}, and repeated updates to a foundation model through fine-tuning \cite{Hu2022LoRA}.


\subsubsection{Reinforcement Learning}

In reinforcement learning, a policy is repeatedly updated from environmental feedback, and its actions shape which states and experiences become available later \cite{Schulman2017PPO}. We train a PPO agent in Hopper-v4 \cite{Schulman2017PPO} under static and dynamic bias. Training has two stages: Clean Learning under the true environmental objective, followed by Bias Exposure. We track true returns and policy updates to test whether performance can deteriorate while optimization remains active.

\paragraph{Static Bias.}
In the static-bias experiment, we introduce a fixed reward bias throughout Bias Exposure. The bias remains constant during exposure and does not depend on the policy state, making it a persistent external error in the learning signal.

Figure~\ref{fig:stable_but_wrong}(a) shows learning under persistent static bias. As the bias continues to affect learning, the agent's return in the true environment declines, while the policy-update-magnitude proxy remains nonzero and we observe no explicit numerical divergence or optimization collapse. These observations indicate continued, non-divergent optimization but do not prove strict convergence or stability.

\paragraph{Dynamic Bias.}
We next examine a bias that directly enters the learning-update dynamics. While keeping the environment, network architecture, and PPO protocol unchanged, we continuously perturb advantage estimation so that an erroneous learning signal directly participates in subsequent policy optimization. Unlike static bias, this bias acts on policy updates and affects later learning states as the policy evolves.

Figure~\ref{fig:stable_but_wrong}(b) shows learning under dynamic bias. Policy updates remain active, and we observe no explicit numerical divergence or optimization collapse. At the same time, returns in the true environment continue to decline and increasingly depart from the true environmental objective.

Both forms of bias reduce true-environment returns while policy updates remain active and non-divergent. Because the experiments do not report a strict convergence statistic, they do not establish strict convergence.


\subsubsection{Supervised Learning}

Unlike RL, supervised learning updates the model from a fixed dataset, without policy--environment interaction. Using CIFAR-10 and a ResNet-18 classifier \cite{Krizhevsky2009CIFAR,He2016ResNet}, we test whether a fixed structural bias and model-driven data feedback can produce the same state.

\paragraph{Static Bias.}
During Bias Exposure, we apply a fixed Structural Patch to the target class while keeping the training protocol, network architecture, and optimization parameters unchanged. The perturbation remains fixed throughout exposure and therefore constitutes persistent static bias in supervised learning.

Figure~\ref{fig:stable_but_wrong}(c) shows the results. As Bias Exposure proceeds, clean test accuracy on the target class declines overall and fluctuates at a lower level, indicating reduced correctness relative to the original task objective. Meanwhile, training loss continues to fall and enters a stable regime, showing that the model continues to optimize its current training objective stably.

The measures move in opposite directions:
\[
\mathrm{Training\ Loss}\downarrow,
\qquad
\mathrm{Target\ Accuracy}\downarrow.
\]

\paragraph{Dynamic Bias.}
We next construct a Prediction-Dependent Label Flip. During Bias Exposure, labels are flipped in a prediction-dependent manner with probability $p=0.5$ for target-class examples only: when the current prediction is correct, the label is changed to the highest-probability incorrect class; when the current prediction is wrong, the label is set to the current prediction, reinforcing the error. This creates a dynamic feedback loop:
\[
\text{Model} \rightarrow \text{Label} \rightarrow \text{Model}.
\]

Figure~\ref{fig:stable_but_wrong}(d) shows the results. Under persistent dynamic bias, clean target-class test accuracy declines overall, although it varies substantially across epochs and seeds. Across the different exposure durations, mean Bias-end accuracy over five seeds ranges from $22.98\%$ to $49.62\%$ and does not decline monotonically with exposure duration. Training loss rises briefly when the bias is introduced, then gradually falls and remains bounded; we observe no evident numerical divergence.


\subsubsection{Large Language Models}

To test whether SBW also occurs during large-language-model fine-tuning, we continually fine-tune Qwen2.5-7B-Instruct with LoRA \cite{Qwen25TechnicalReport,Hu2022LoRA} on GSM8K multi-step math problems \cite{Cobbe2021GSM8K}. We divide GSM8K into Target and Non-target subsets. Target Accuracy measures final-answer correctness against the original answers; Training Loss measures optimization of the current objective. The static and dynamic conditions use fixed and feedback-coupled erroneous supervision, respectively.

\paragraph{Static Bias.}
Under static bias, a fixed erroneous-supervision rule is applied to the Target subset throughout Bias Exposure; supervision on all other examples remains correct.

Figure~\ref{fig:stable_but_wrong}(e) shows Qwen2.5-7B-Instruct under static bias. The blue curve represents Training Loss, measured with a consistent sliding-window statistic to characterize optimization of the current training objective. The red curve represents Target Accuracy, which measures correctness relative to the original correct reasoning objective.

As Bias Exposure continues, Training Loss falls and enters a stable regime while Target Accuracy declines, showing that optimizing the current supervision does not imply improved performance on the original task.

\paragraph{Dynamic Bias.}
Under dynamic bias, the model's generated final answers are repeatedly used to refresh the supervision state after an initial Bias Exposure. The model state therefore influences the learning signal at the next stage, allowing bias to evolve during learning.

Figure~\ref{fig:stable_but_wrong}(f) shows the dynamic-bias results. We use the same fixed-window Training Loss statistic as in the static condition. During Bias Exposure, Training Loss falls and approaches a stable regime, while Target Accuracy declines relative to the original answers.

The two metrics move in opposite directions:
\[
\mathrm{Training\ Loss}\downarrow,
\qquad
\mathrm{Target\ Accuracy}\downarrow.
\]


\subsubsection{Cross-Paradigm Evidence for Stable but Wrong}

Across the three paradigms, optimization of the current signal remains active or enters a stable regime while performance relative to the external objective declines or remains displaced. In RL, policy updates remain active without explicit numerical divergence or collapse as true returns deteriorate. In SL, Training Loss falls under static bias and, under dynamic bias, falls after a brief increase; target-class Bias-end accuracy in Fig.~\ref{fig:stable_but_wrong}(c,d) is below the end of the Clean stage in both cases. During continual fine-tuning of Qwen2.5-7B-Instruct, Training Loss falls and stabilizes while Target Accuracy declines. This pattern occurs under both static and dynamic bias and across different architectures and optimization mechanisms.

\begin{quote}
\textbf{Convergence characterizes learning stability, not learning correctness.
Under persistent bias in the learning signal, a system can continue to optimize and stabilize
while remaining systematically displaced from the true objective.}
\end{quote}

\[
\boxed{
\mathrm{Learning\ Stability}
\;\not\Rightarrow\;
\mathrm{Learning\ Correctness}
}.
\]

\subsection{Challenging and Correcting Stable but Wrong}
\label{sec:challenge-correction}

The preceding experiments show that a model can retain an erroneous judgment as learning stabilizes. We next test whether external evidence can identify and change a specified judgment.

The experiments assume ground truth independent of the learning process and a prespecified judgment to test. We compare that judgment with the external truth, then retrain the model on evidence relevant to the error. They provide a controlled test of whether \textbf{an established error can be identified and corrected under these conditions}.

\paragraph{Controlled scientific discovery.}
We first test this in a five-dimensional scientific-discovery task with a known true utility function. After autonomous search, the model stably ranks a broad, accessible region A above a narrow, remote region B. External ground truth, however, shows that the mean outcome in the neighborhood of B is higher than in A. During the Challenge stage, we do not update the model or add true outcomes to its history; we only compare its predicted ranking of the prespecified regions A and B with the ranking given by the external ground truth. In all 19 runs included in the analysis, the model predicts $A>B$ while the true ranking is $B>A$, confirming the erroneous ranking in all 19 prescreened runs. As a control, we construct a matched environment with the same historical sample locations and observations but a true landscape in which A is indeed better than B. In that environment, the model's ranking agrees with the true ranking in all 19 runs, with no correct judgment misclassified as an error.

During Correction, we start independently from each frozen history, add true observations at a small number of new locations, retrain the model, and compare the predicted A--B ranking again. Adding 1, 2, 5, or 10 random true observations corrects the erroneous ranking in none of the 19 runs. In contrast, adding observations from the B neighborhood produces an overall increase in corrections with more evidence: 13/19 runs are corrected with one observation, 16/19 with two, and 19/19 with either five or ten. When using the experimenter-known center of B as an upper-bound condition, one observation is sufficient to change the predicted ranking to $B>A$ in all 19 runs. In the matched correct environment, none of the tested interventions flips the originally correct ranking $A>B$. These results indicate that correction does not come from adding data in general; it depends on whether the new evidence directly challenges the original erroneous judgment.

\paragraph{Image classification.}
We next test whether an established erroneous feature dependence can be identified and reduced in CIFAR-10. During Bias Exposure, automobile images continuously receive a $4\times4$ Structural Patch unrelated to class semantics, giving the model an opportunity to use the marker as a shortcut. Because the marker is added by the experimenter, its irrelevance to the true class provides external ground truth. During Challenge, on an independent set of non-automobile test images, we compare model predictions with and without the marker and define shortcut sensitivity as the proportion of predictions that change. The mean shortcut sensitivity is 0.335 for the Wrong checkpoint after bias exposure, compared with 0.012 for the matched Clean checkpoint, indicating that the model has developed a clear dependence on the erroneous marker.

We then continue training with relevant counterexamples that directly break the association “marker implies automobile”: we add the same marker to images whose true class is not automobile while retaining their true labels; the other half consists of clean automobile images. The matched control uses exactly the same images and labels but omits the marker from non-automobile images, isolating evidence against the erroneous association from general data addition. At the prespecified budget $k=50$, relevant counterexamples reduce shortcut sensitivity by 0.234, compared with a reduction of 0.087 for the matched cue-absent control. The seed-level paired advantage is 0.146, with the same direction across all five source seeds (one-sided exact Wilcoxon $p=0.03125$; seed-level bootstrap 95\% CI $[0.120,0.177]$). After the relevant-counterexample intervention, shortcut sensitivity falls from 0.335 to about 0.101, while mean automobile-class accuracy rises from 48.3\% to 80.5\%. The same updates to the paired Correct checkpoint produce no evident erroneous dependence or overall performance damage.

\begin{figure}[H]
\centering
\includegraphics[width=0.96\linewidth]{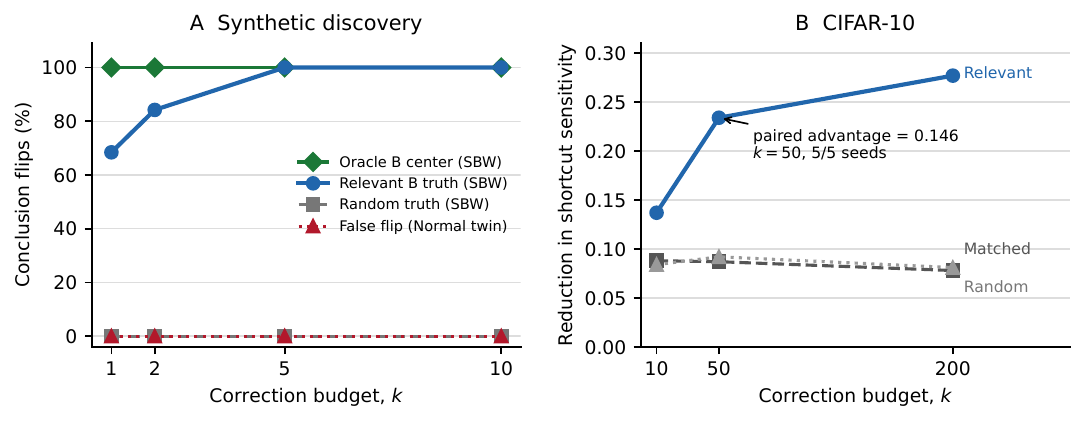}
\caption{\textbf{Targeted correction after challenge by external ground truth.} (A) Controlled scientific discovery: the y-axis shows the fraction of 19 runs in which the erroneous A--B ranking is flipped. B center is an upper-bound condition using the experimenter-known location; B neighborhood provides relevant observations; random observations are the control. No arm produces an erroneous flip in the Normal twin environment. (B) CIFAR-10: reduction in shortcut sensitivity under different update budgets; marked counterexamples and matched cue-absent controls use the same base images and labels. The y-axis metrics differ and must not be compared across tasks. Pre-update Challenge results are reported in Table~\ref{tab:challenge_correction_main}.}
\label{fig:challenge_correction}
\end{figure}

Figure~\ref{fig:challenge_correction} shows results across correction budgets for both experiments. Table~\ref{tab:challenge_correction_main} summarizes the Challenge judgments before model updates and the Correction results after updates; full methods and controls are provided in Appendix~\ref{app:challenge-correction}.

\begin{table}[H]
\centering
\small
\caption{Challenge and correction results in two controlled experiments. The synthetic experiment uses 19 prescreened paired runs; the primary CIFAR-10 comparison is based on five source seeds.}
\label{tab:challenge_correction_main}
\begin{tabularx}{\linewidth}{p{0.16\linewidth}X X}
\hline
System & Ground-truth challenge (before model update) & Targeted correction and control on correct states \\
\hline
Controlled scientific discovery & In the SBW environment, the model ranks A above B in 19/19 runs, opposite to the true ranking; in the Normal environment, the two rankings agree in 19/19 runs & One true observation at the B center ($k=1$) changes the SBW ranking in 19/19 runs; five true observations from the B neighborhood ($k=5$) change it in 19/19 runs, whereas 5 random observations yield 0/19; all Normal arms together show 0/19 erroneous flips \\
CIFAR-10 & Mean $S_{\mathrm{shortcut}}=0.335$ for the Wrong checkpoint versus $0.012$ for the paired Correct checkpoint & At $k=50$, relevant counterexamples reduce $S$ by $0.234$, matched controls by $0.087$, for a paired advantage of $0.146$; updates to the Correct checkpoint do not trigger the prespecified safety failure \\
\hline
\end{tabularx}
\end{table}

The two experiments show that independent evidence can expose and reduce a prespecified error under controlled conditions. In scientific discovery, a small number of observations from an overlooked region can change the model's ranking. In image classification, counterexamples that break the erroneous association reduce shortcut sensitivity more than matched controls. \textbf{These results establish detectability and correctability only when external ground truth and the judgment to be tested are available.} Future work can examine whether local validation, counterfactual probes, or active evidence acquisition can identify possible SBW states when complete ground truth is unavailable.

\section{Conclusion}

Stable but Wrong (SBW) describes a learning state in which persistent bias keeps the model's knowledge or decisions displaced from an external objective even as optimization proceeds and stabilizes. It distinguishes stability of the learning process from correctness of its outcome.

Experiments in reinforcement learning, supervised learning, and large language models show this pattern under different architectures and bias mechanisms. In reinforcement learning, policy updates remain active without evident numerical instability as true environmental returns decline. In supervised learning, training loss falls under static bias and, after a brief increase, under dynamic bias, while target-class accuracy at Bias-end is below the end of the Clean stage. During continual fine-tuning, training loss stabilizes while final-answer accuracy on the original task declines. These results show that optimization can remain stable relative to its current signal even as performance on the external objective worsens.

Two controlled challenge--correction experiments show that targeted evidence can change an established error. In scientific discovery, observations from an overlooked high-value region change the model's ranking, whereas random observations do not have the same effect. In image classification, counterexamples that break an erroneous marker association reduce shortcut sensitivity more than matched controls. We observe no evident harm to paired correct models under the tested protocol.

These experiments establish detectability and correctability only when external ground truth and the judgment to be tested are available; they do not provide a general method for discovering unknown errors. Future work should test whether local validation, counterfactual probes, independent data sources, or active evidence acquisition can identify errors when complete ground truth is unavailable.

Falling loss, convergence, and stable predictions show adaptation to the current learning signal. When that signal is persistently biased, these measures alone cannot establish that the model's knowledge agrees with the true objective. Reliability assessments for systems that learn over time therefore require independent evaluation of whether their conclusions withstand external evidence.

\begin{center}
\fbox{\parbox{0.72\linewidth}{
\centering
\textbf{Learning Stability $\neq$ Learning Correctness.}\\[0.4em]
Stability shows adaptation to the current signal;\\
independent evaluation is needed to assess correctness.
}}
\end{center}

\clearpage
\begin{appendices}

\section[Cross-Environment Validation in Reinforcement Learning]{Cross-Environment Validation in\\ Reinforcement Learning}
\label{app:rl_cross_environment}

We examine Stable but Wrong (SBW) under controlled reinforcement-learning conditions using Progressive Bias Exposure experiments on MuJoCo continuous-control tasks. The experiments cover static and dynamic bias across Hopper-v4 and Walker2d-v4, multiple bias strengths, and Bias Exposure durations.

\subsection*{Experimental setup}

\textbf{Environments and algorithm.} We use the Gymnasium/MuJoCo environments Hopper-v4 and Walker2d-v4 and PPO from Stable-Baselines3 (\texttt{MlpPolicy}; default hyperparameters: learning rate $3\times10^{-4}$, $n_{\mathrm{steps}}=2048$, batch size 64, $n_{\mathrm{epochs}}=10$, $\gamma=0.99$, and GAE $\lambda=0.95$), with $4\times$\texttt{DummyVecEnv} parallel environments. Each configuration uses five random seeds ($0$--$4$). We evaluate every $10\mathrm{k}$ steps using a deterministic policy over five episodes in the clean environment and record the true return.

\textbf{Two-stage protocol.} Training has Clean and Bias Exposure stages. During Clean ($0$--$1.0\mathrm{M}$ steps), the agent trains under the true reward to establish a baseline. During Bias Exposure ($1.0\mathrm{M}$--$1.0\mathrm{M}+E$), bias is continuously introduced, with exposure duration $E\in\{0.25,0.5,0.75,1.0\}\mathrm{M}$. The Clean reference is the mean true return over the final $100\mathrm{k}$ steps of Clean; the Bias-end return is the mean over the final $100\mathrm{k}$ steps of Bias Exposure.

\textbf{Static bias.} During Bias Exposure, we add a fixed shift to each reward:
\[
r_{\mathrm{obs}}=r_{\mathrm{true}}+c,\qquad c=-1.0.
\]
The shift is independent of policy state, does not change the sign of the advantage, and does not enter a policy-dependent feedback-selection process. Evaluation always uses the clean reward, separating contamination of the training signal from assessment of true performance. The static-bias experiment on Hopper-v4 covers the full grid of exposure durations.

\textbf{Dynamic bias.} The environment reward remains the true reward. During Bias Exposure, we flip the sign of rollout-buffer advantages with probability $p$ (advantage sign flip). The perturbation directly enters policy-gradient updates and creates the feedback-coupled loop
\[
\text{Policy}\rightarrow\text{Trajectory}\rightarrow\text{Advantage}\rightarrow\text{Update}\rightarrow\text{Policy}.
\]
On Hopper-v4 and Walker2d-v4, we cover all combinations of bias strength $p\in\{0.3,0.5,0.7\}$ and exposure duration, for a total of $2\times3\times4\times5=120$ independent runs.

\textbf{Evaluation metrics.} True Return measures performance displacement relative to the true environmental objective. We monitor the training process using a policy-update-magnitude proxy and training logs for numerical divergence or training collapse. The proxy describes only whether optimization remains active and non-divergent; it is not evidence of strict convergence.

\subsection*{Results: Stable but Wrong under persistent bias}

Tables~\ref{tab:rl_sbw_static} and~\ref{tab:rl_sbw_dynamic} report performance relative to Clean at the end of Bias Exposure. Under static bias, Hopper returns fall by about $30\%$--$42\%$ across exposure durations, while policy updates remain active and training shows no explicit numerical divergence or collapse. Performance on the true objective can therefore deteriorate without an explicit training failure.

\begin{table}[ht]
\centering
\caption{Evidence of SBW under static bias (reward shift $c=-1.0$) on Hopper-v4 (five seeds, mean $\pm$ standard deviation). $R_{\mathrm{bias-end}}$ is the true return at the end of Bias Exposure.}
\label{tab:rl_sbw_static}
\begin{tabular}{c|c|c|c}
\hline
Exposure duration & Clean reference & Bias-end return & Relative decrease \\
\hline
$0.25\mathrm{M}$ & $3193\pm224$ & $2046\pm230$ & $35.9\%$ \\
$0.50\mathrm{M}$ & $3193\pm224$ & $1990\pm295$ & $37.7\%$ \\
$0.75\mathrm{M}$ & $3193\pm224$ & $2250\pm698$ & $29.6\%$ \\
$1.00\mathrm{M}$ & $3193\pm224$ & $1862\pm459$ & $41.7\%$ \\
\hline
\end{tabular}
\end{table}

Under dynamic bias, the effect depends strongly on $p$. At $p=0.3$, performance changes are weak or inconsistent and do not produce a clear SBW pattern. At $p=0.5$, both environments show moderate to substantial declines. At $p=0.7$, relative decreases exceed $90\%$ under long exposure in both Hopper and Walker. Thus, small perturbations do not necessarily produce this pattern; sufficiently strong bias in the update dynamics can substantially reduce true-task performance without explicit numerical divergence or collapse.

\begin{table}[ht]
\centering
\caption{Performance displacement from Clean at the end of Bias Exposure under dynamic bias (advantage flip; means across five seeds). Relative decreases are shown in parentheses.}
\label{tab:rl_sbw_dynamic}
\begin{tabular}{c|c|c|c|c}
\hline
Environment & $p$ & Exposure $0.25\mathrm{M}$ & Exposure $0.50\mathrm{M}$ & Exposure $1.00\mathrm{M}$ \\
\hline
Hopper & $0.3$ & $2868$ ($8.9\%$) & $3191$ ($-1.3\%$) & $2496$ ($20.7\%$) \\
Hopper & $0.5$ & $1493$ ($52.6\%$) & $1388$ ($55.9\%$) & $494$ ($84.3\%$) \\
Hopper & $0.7$ & $336$ ($89.3\%$) & $151$ ($95.2\%$) & $49$ ($98.4\%$) \\
Walker & $0.3$ & $2417$ ($-5.5\%$) & $2489$ ($-8.6\%$) & $2623$ ($-14.5\%$) \\
Walker & $0.5$ & $2053$ ($10.4\%$) & $1593$ ($30.4\%$) & $1290$ ($43.7\%$) \\
Walker & $0.7$ & $1148$ ($49.9\%$) & $319$ ($86.1\%$) & $44$ ($98.1\%$) \\
\hline
\end{tabular}
\end{table}

\subsection*{Overall discussion}

The RL experiments show that both a fixed reward shift and dynamic advantage flips can reduce true-task performance while policy updates remain active and training shows no explicit numerical divergence or collapse. Dynamic-bias effects depend on bias strength and exposure duration. Because we do not report a strict convergence statistic, these results do not establish strict stable convergence.

\section{Validation with Qwen2.5-7B-Instruct}
\label{app:qwen_validation}

We also evaluate Qwen2.5-7B-Instruct \cite{Qwen25TechnicalReport} on GSM8K multi-step math problems \cite{Cobbe2021GSM8K}. The protocol alters final-answer supervision rather than systematically rewriting intermediate reasoning. The results therefore characterize outcome-level bias during continual fine-tuning; they do not establish that the model's internal reasoning structure has changed.

\subsection*{Experimental setup}

We continually fine-tune Qwen2.5-7B-Instruct with LoRA on the GSM8K \texttt{main/test} split (rank$=16$, applied to query/value projection layers; $\alpha=32$, dropout$=0.05$) \cite{Qwen25TechnicalReport,Cobbe2021GSM8K,Hu2022LoRA}. Training uses completion-only loss, maximum sequence length 1024, learning rate $2\times10^{-4}$, batch size 1, gradient accumulation 8, a constant scheduler, and AdamW. Unless otherwise noted, we use five random seeds. The available engineering records do not include immutable hashes for the model weights or dataset revision. We divide the GSM8K test set using a \textbf{prespecified deterministic question-text rule}: after lowercasing each question, we assign it to Target if it contains any of the strings \texttt{percent}, \texttt{percentage}, \texttt{\%}, \texttt{ratio}, \texttt{fraction}, \texttt{out of}, or \texttt{per cent}; all other questions are assigned to Non-target. This rule yields:
\begin{itemize}
    \item \textbf{Target subset} (207 questions): measures final-answer accuracy on problem types exposed to bias;
    \item \textbf{Non-target subset} (1112 questions): contains the remaining test questions and is used to assess broad performance degradation.
\end{itemize}
This deterministic rule can be applied directly to the public GSM8K \texttt{main/test} split to obtain the reported 207/1112 partition. The complete Target list was not included with the current manuscript or arXiv source files.

The experiment uses a two-stage protocol:
\[
\mathrm{Clean}\rightarrow \mathrm{Bias\ Exposure}.
\]
During Clean, the model is fine-tuned on the original GSM8K data for 500 steps to establish baseline reasoning ability. During Bias Exposure, systematic erroneous supervision is introduced.

\subsection*{Static-bias mechanism}

For each Target training example, $y_{\mathrm{true}}$ is defined as the first numerical value after \texttt{\#\#\#\#} in the original GSM8K solution. If this marker is absent, we use the last numerical value in the solution text. Examples for which no number can be parsed are excluded. During Bias Exposure, we apply the fixed bias rule
\[
y_{\mathrm{bias}} = 1.25 \times y_{\mathrm{true}},
\]
which remains unchanged throughout the Exposure stage. To construct a biased completion, we retain all original solution text before \texttt{\#\#\#\#}, append a fixed scaling rule and calculation, and replace the final \texttt{\#\#\#\#} answer with the formatted $y_{\mathrm{bias}}$ (integers remain integers; other values retain at most six decimal places with trailing zeros removed). The intervention therefore changes the final answer and adds a final scaling step; it does not systematically rewrite the original intermediate reasoning. Stage 1 uses all 1164 Target training examples and 1746 Non-target examples sampled by random seed, forming a fixed mixed training set of 2910 examples (40\% Target-bias). Exposure durations are
\[
T \in \{1000, 3000, 5000, 8000\},
\]
with five random seeds per condition.

\subsection*{Dynamic-bias mechanism (Exposure only)}

Dynamic bias uses a feedback-coupled supervision-memory (\texttt{memory\_feedback}) protocol. It begins with 200 steps of seed Bias Exposure. Every 500 steps thereafter, the model generates answers greedily for all 1164 Target training questions using a Qwen chat prompt containing only the question (\texttt{do\_sample=false}, temperature$=0$, top-p$=1$, top-k$=0$, maximum 512 new tokens). We parse the generated final numerical values using the same \texttt{\#\#\#\#}/last-number rule as in the static experiment. The generated text itself is \textbf{not} used directly as training supervision; it is used only to select the next supervision state between the existing clean completion and the fixed biased completion. For each question, the supervision state is retained with probability 0.8. In the remaining 0.2 refresh probability, if the prediction is more than 0.5 away from both the true answer and biased answer, the previous state is retained; otherwise, the closer side is selected, with ties retaining the previous state. The training set is then rebuilt using the complete clean or fixed biased completions. Each seed stops early when bias consistency is stable and Target Accuracy changes by no more than 0.10 relative to the end of seed-bias exposure: bias-consistency changes by less than 0.03 for two consecutive evaluations. Total exposure steps are therefore not required to align. The main text reports checkpoints shared by all five seeds:
\[
\text{global step} \in \{200, 700, 1200, 1700\}.
\]

\subsection*{Evaluation metrics}

We assess the core SBW evidence along three dimensions:
\begin{itemize}
    \item \textbf{Stable}: whether Training Loss decreases and enters a stable regime during Bias Exposure. To make different Exposure checkpoints comparable, Training Loss is defined consistently as the mean loss over the $W{=}100$ training steps before the corresponding checkpoint. Step~0 uses the mean loss over the first 100 steps after Bias Exposure begins. Specifically, checkpoint $C{=}0$ uses the arithmetic mean of recorded losses at global steps $[1,100]$; checkpoint $C{>}0$ uses the mean over $[C{-}99,\,C]$ (HF Trainer, $\texttt{logging\_steps}{=}20$).
    \item \textbf{Wrong}: whether final-answer accuracy on the Target subset declines systematically;
    \item \textbf{Control}: whether Non-target Accuracy remains relatively high, ruling out overall collapse.
\end{itemize}

\subsection*{Static-bias results}

At the end of Clean, Target Accuracy is $67.05\%\pm2.35\%$ and Non-target Accuracy is $76.98\%\pm1.26\%$ (five seeds, mean $\pm$ standard deviation).

Table~\ref{tab:qwen_static_sbw} reports the core SBW evidence under static bias. Using the consistent $W{=}100$ window-mean definition, Training Loss declines continuously from the start of Bias Exposure ($0.237\pm0.034$) and stabilizes at $0.007$--$0.011$ later ($0.237 \rightarrow 0.055 \rightarrow 0.011 \rightarrow 0.008 \rightarrow 0.007$), indicating continued optimization of the current supervision. Target Accuracy falls from $11.30\%\pm4.54\%$ at $T{=}1000$ to about $6\%$ for $T\ge 3000$. Non-target Accuracy falls from $76.98\%$ to about $70\%$, remaining substantially higher than Target. This indicates that final-answer displacement is concentrated mainly in the contaminated problem types rather than reflecting overall training collapse.

\begin{table}[ht]
\centering
\caption{Core evidence of SBW under static bias in Qwen2.5-7B-Instruct (five seeds, mean $\pm$ standard deviation). Training Loss is the $W{=}100$ window mean (see the evaluation-metrics description).}
\label{tab:qwen_static_sbw}
\begin{tabular}{c|c|c|c}
\hline
Exposure duration & Training Loss & Target Accuracy & Non-target Accuracy \\
\hline
Clean              & --                & $67.05\pm2.35\%$ & $76.98\pm1.26\%$ \\
Exposure start (0) & $0.237\pm0.034$   & $67.05\pm2.35\%$ & $76.98\pm1.26\%$ \\
1000               & $0.055\pm0.002$   & $11.30\pm4.54\%$ & $74.53\pm1.53\%$ \\
3000               & $0.011\pm0.001$   & $6.47\pm0.99\%$  & $68.42\pm2.31\%$ \\
5000               & $0.008\pm0.001$   & $5.80\pm1.01\%$  & $69.86\pm1.67\%$ \\
8000               & $0.007\pm0.001$   & $6.38\pm1.54\%$  & $69.62\pm3.13\%$ \\
\hline
\end{tabular}
\end{table}

\subsection*{Dynamic-bias results (Exposure)}

Table~\ref{tab:qwen_dynamic_sbw} reports the SBW evidence under dynamic bias at checkpoints aligned across all five seeds. Using the same $W{=}100$ window mean, Training Loss falls from the start of Bias Exposure ($0.246\pm0.030$) and approaches a stable regime ($0.246 \rightarrow 0.133 \rightarrow 0.065 \rightarrow 0.037 \rightarrow 0.022$). Target Accuracy declines from $31.01\%\pm6.82\%$ at step $=200$ to $13.43\%\pm3.18\%$ at step $=1700$. Non-target Accuracy remains around $70\%$--$74\%$, indicating that the Wrong effect remains concentrated mainly in the Target subset, although dynamic coupling has a more visible spillover effect on Non-target than in the static condition.

\begin{table}[ht]
\centering
\caption{Core evidence of SBW under dynamic bias in Qwen2.5-7B-Instruct (five seeds, aligned checkpoints, mean $\pm$ standard deviation). Training Loss is the $W{=}100$ window mean, using the same protocol as in the static condition.}
\label{tab:qwen_dynamic_sbw}
\begin{tabular}{c|c|c|c}
\hline
Global Step & Training Loss & Target Accuracy & Non-target Accuracy \\
\hline
0 (exposure start) & $0.246\pm0.030$ & $67.05\pm2.35\%$ & $76.98\pm1.26\%$ \\
200                & $0.133\pm0.001$ & $31.01\pm6.82\%$ & $70.85\pm3.03\%$ \\
700                & $0.065\pm0.001$ & $19.13\pm8.15\%$ & $74.03\pm0.92\%$ \\
1200               & $0.037\pm0.002$ & $15.75\pm5.12\%$ & $73.38\pm1.25\%$ \\
1700               & $0.022\pm0.001$ & $13.43\pm3.18\%$ & $72.23\pm1.37\%$ \\
\hline
\end{tabular}
\end{table}

\subsection*{Overall discussion}

The Qwen2.5-7B-Instruct experiment shows that the SBW risk pattern can occur when a modern instruction-tuned LLM learns multi-step math problems. Under both static and dynamic supervision, Training Loss falls and stabilizes during Bias Exposure while final-answer accuracy on Target declines (Training Loss$\downarrow$/converges $+$ Target Accuracy$\downarrow$). Under dynamic Exposure, Target accuracy continues to deteriorate and Non-target performance also shows more visible spillover effects. This supports a separation between optimization of the current supervision and final-answer correctness on the original task; it does not, by itself, establish that the model's internal multi-step reasoning structure has changed.

\section{Medical-Imaging Motivation Example}
\label{app:medical_motivation}

This experiment is included only as a motivational example to illustrate Stable but Wrong; it is not part of the paper's main empirical evaluation.

\subsection{Dataset}

We use the public PadChest chest-radiograph dataset for this motivational example \cite{Bustos2020PadChest}. The dataset provides standardized acquisition metadata, including projection-view fields, allowing us to construct controlled acquisition bias while retaining real clinical images. Chest-radiograph data-processing tools are available in TorchXRayVision \cite{Cohen2022TorchXRayVision}.

The experiment uses a filtered subset of 433 images. The positive class is Cardiomegaly (218 images); the negative class contains images labeled strictly as \texttt{['normal']} (215 images). There are 221 AP-view images and 212 PA-view images.

\subsection{Construction of the systematic bias}

We use controlled sample selection to create perfect collinearity between disease labels and projection view in the training and validation data.

All Cardiomegaly examples come from the AP view, and all Normal examples come from the PA view. The AP/PA projection views and radiographs themselves come from real clinical acquisition data; the perfect collinearity is experimentally constructed through sample selection to isolate the effect of acquisition domain as a proxy feature. The test set additionally retains real examples that break the association (Cardiomegaly+PA and Normal+AP). We do not synthesize images, modify pixels, redefine labels, or intervene manually at the feature level. Splits are fixed at the patient level; the split counts are reported in the table, but the patient-level split file was not included with the current manuscript or arXiv source files.

\subsection{Split statistics}

The dataset is divided as follows:

\begin{table}[ht]
\centering
\begin{tabular}{lccc}
\hline
Class and view & Training & Validation & Test \\
\hline
Cardiomegaly + AP & 143 & 21 & 29 \\
Normal + PA & 132 & 27 & 28 \\
Cardiomegaly + PA & 0 & 0 & 25 \\
Normal + AP & 0 & 0 & 28 \\
\hline
Total & 275 & 48 & 110 \\
\hline
\end{tabular}
\caption{Dataset splits. The final two groups (Cardiomegaly+PA and Normal+AP) do not appear in training or validation and are used to test whether the model relies on an acquisition-view shortcut.}
\end{table}

\subsection{Model and training}

We use an ImageNet-1K-pretrained ResNet-18 backbone \cite{He2016ResNet} and replace the final fully connected layer with a single logit for fine-tuning using BCEWithLogitsLoss. Single-channel X-rays are copied to three channels and normalized using ImageNet means and standard deviations. We use Adam with learning rate $1\times10^{-4}$ and batch size 32, train for 20 epochs, and select the best checkpoint by validation AUC. We do not use chest-radiograph-specific pretrained weights, retaining a standard supervised-learning baseline.

\subsection{Results}

The model reaches 100.0\% training accuracy and 97.9\% validation accuracy. On the test set, mean accuracy on aligned examples (Cardiomegaly+AP and Normal+PA) is 98.2\%, compared with 39.6\% on examples that break the collinearity (Cardiomegaly+PA and Normal+AP). Performance therefore drops sharply when the shortcut association between disease labels and projection view is removed, despite successful optimization.

\begin{table}[ht]
\centering
\begin{tabular}{lcc}
\hline
Group & Number of examples & Accuracy \\
\hline
Cardiomegaly + AP (aligned) & 29 & 100.0\% \\
Normal + PA (aligned) & 28 & 96.4\% \\
Cardiomegaly + PA (association broken) & 25 & 40.0\% \\
Normal + AP (association broken) & 28 & 39.3\% \\
\hline
\end{tabular}
\caption{Classification accuracy on biased test examples and examples that break the bias.}
\end{table}

\subsection{Discussion}

This example uses real clinical radiographs and their original AP/PA acquisition attributes. The perfect association between labels and acquisition domain is created through controlled sample selection; we do not claim that it occurs naturally in the original PadChest distribution. The results show that stable optimization can reinforce acquisition features unrelated to disease semantics when these attributes are made a strong proxy for the label. Figure~\ref{fig:intro_medical_sbw} summarizes the example, which serves as motivation rather than as a benchmark experiment.

\section{Motivational Example of Stable but Wrong in a Controlled Synthetic Autonomous-Discovery System}
\label{app:intro_autolab_sbw}

This appendix describes the synthetic autonomous-discovery example in the Introduction (Fig.~\ref{fig:intro_autolab_sbw}); methods and results for the paired Challenge--Correction experiment appear in Appendix~\ref{app:challenge-correction}. Neither experiment uses A-Lab data, hardware, or outcomes, or reproduces A-Lab. The Introduction example is a motivation, not a primary experiment, a Bayesian-optimization benchmark, or an evaluation of feedback-coupling theory.

\subsection{Purpose and scope}

In autonomous scientific discovery, the learner helps determine which experiments are run next, forming a loop of “experiment selection $\rightarrow$ feedback $\rightarrow$ model update $\rightarrow$ reselection.” We use this controlled system to ask:

\emph{Can a learning system operate stably, with stable predictive dynamics, while forming an incorrect scientific belief?}

We do \textbf{not} define Wrong as search failure, insufficient exploration, or failure of the acquisition function. Wrong means a \textbf{mismatch between the learned scientific belief and the oracle ground truth}.

\subsection{Experimental setup}

We simulate autonomous materials discovery with Bayesian optimization in a five-dimensional continuous candidate space, $x\in[0,1]^5$. Each observation updates a Gaussian-process surrogate implemented with BoTorch \texttt{SingleTaskGP}. The learner selects the next location using Log Expected Improvement (\texttt{qLogExpectedImprovement}, sequential $q=1$, Kriging Believer). Candidate rewards come from a synthetic multimodal objective with noisy observations; the learner cannot access the Oracle during the run.

\begin{itemize}
  \item \textbf{Oracle evaluation:} a dense offline scan of the deterministic objective $f(x)\equiv R_{\mathrm{det}}(x)$, used only for post hoc evaluation.
  \item \textbf{Reference run:} the \texttt{baseline} condition, with $n_{\mathrm{init}}{=}32$, $T{=}30$ generations, and 10 experiment points per generation.
  \item \textbf{Implementation records:} configuration files and intermediate results for this example were maintained as internal experimental records and were not included with the current manuscript or arXiv source files.
\end{itemize}

The procedure is to fit the GP, maximize the acquisition function, observe the selected point, and update the dataset. Figure~\ref{fig:intro_autolab_sbw} measures Stable and Wrong in separate panels; it does not directly visualize the five-dimensional search space.

\subsection{Results}

\textbf{Stable (Panel~A).}
Panel~A reports the Gaussian-process root mean squared prediction error (RMSE) on a fixed validation pool. This pool is not used to train the GP and measures whether the surrogate model's predictive dynamics remain stable. Across five random seeds, RMSE falls during early generations and then enters a bounded stable regime, indicating that the surrogate model retains stable predictive ability without evident prediction degradation.

\textbf{Wrong (Panel~B).}
Panel~B compares the final \emph{learned scientific belief} of the system with the \emph{ground-truth optimum}. In the example run (\texttt{baseline}, seed$=0$), they are systematically inconsistent: the scientific state considered optimal by the model is displaced from the true optimum revealed by the Oracle. The two-dimensional landscape and colored basin in Panel~B are schematic visualizations; they are \textbf{not} metric projections of the five-dimensional coordinates or search trajectory.

The true optimum remains unchanged throughout the run. The system ends in an SBW state because its learned scientific belief disagrees with that truth: \emph{it learns stably, yet forms an incorrect belief.}

\subsection{Measurement definitions}

\paragraph{Stable: GP prediction RMSE.}
\textbf{Stable} refers to the surrogate model's \emph{stable predictive dynamics}: prediction error on held-out candidates remains bounded and enters a stable regime. It does not refer to whether the true global optimum has been found. We compute the following quantity on a fixed validation pool $\mathcal{V}=\{x_i^{(v)}\}_{i=1}^{N}$ ($N{=}4096$, Sobol, seed 424242, and \textbf{not used for GP training}):
\begin{equation}
  \mathrm{RMSE}_t =
  \sqrt{
    \frac{1}{N}
    \sum_{i=1}^{N}
    \bigl(
      f(x_i^{(v)}) - \hat{y}_t(x_i^{(v)})
    \bigr)^2
  },
  \label{eq:autolab-rmse}
\end{equation}
where $f(x)$ is the deterministic Oracle objective and $\hat{y}_t(x)$ is the GP posterior mean at generation $t$. Across five random seeds (\texttt{baseline}), $\mathrm{RMSE}_t$ falls early and stabilizes near $0.06$--$0.07$ after approximately generation 15.

\paragraph{Wrong: learned belief versus ground truth.}
\textbf{Wrong} is defined as a systematic discrepancy between learned belief and oracle truth. At the final generation, we extract the following from the recorded candidate pool $\mathcal{P}$ (2048 Sobol points):
\begin{equation}
  x_{\mathrm{belief}}
  =
  \arg\max_{x \in \mathcal{P}}
  \mu_{\mathrm{GP}}(x),
  \label{eq:autolab-belief}
\end{equation}
that is, “where the model believes the optimum lies.” The ground-truth optimum is determined independently of the learner through an offline Oracle scan:
\begin{equation}
  x_{\mathrm{true}}
  =
  \arg\max_{x} f(x),
  \label{eq:autolab-truth}
\end{equation}
yielding $f(x_{\mathrm{true}})\approx1.0$ (\texttt{oracle\_eval.py}). The SBW condition is $x_{\mathrm{belief}}\neq x_{\mathrm{true}}$ (example run: \texttt{baseline}, seed$=0$).

\subsection{Correspondence with the figure}

Figure~\ref{fig:intro_autolab_sbw} contains two panels that correspond directly to the measurements above.

\paragraph{Panel A.}
\begin{itemize}
  \item \textbf{Quantity plotted:} $\mathrm{RMSE}_t$ (Eq.~\eqref{eq:autolab-rmse}), mean $\pm$ standard deviation across five seeds.
  \item \textbf{Data file:} \texttt{results/autolab\_gp\_rmse.csv}.
  \item \textbf{Interpretation:} model predictions remain stable; this does not indicate improved discovery performance or successful search.
\end{itemize}

\paragraph{Panel B.}
\begin{itemize}
  \item \textbf{Semantic quantities:} $x_{\mathrm{belief}}$ (Eq.~\eqref{eq:autolab-belief}) and $x_{\mathrm{true}}$ (Eq.~\eqref{eq:autolab-truth}).
  \item \textbf{Annotation:} “Learned belief $\neq$ Ground truth”.
  \item \textbf{Interpretation:} the model's scientific belief disagrees with the true regularity; this does not indicate failure of the BO search.
\end{itemize}

The two-dimensional landscape and basin in Panel~B are only schematic visualizations. Axis positions do not encode physical distance or regret; the figure deliberately omits search trajectories, distance annotations, and regret values.

\section{Persistent Bias in Supervised Learning}
\label{app:supervised_learning}

We test whether persistent bias can produce Stable but Wrong (SBW) in supervised learning using controlled CIFAR-10 image-classification experiments. The experiments cover static bias, dynamic bias, and self-induced data drift. We also evaluate Structural Patch and Self-Induced Data Drift on Tiny-ImageNet-200 to assess cross-dataset replication.

\subsection*{Experimental setup}

\textbf{Model and data.} We use CIFAR-10 and a ResNet-18 classifier with a CIFAR-style stem \cite{Krizhevsky2009CIFAR,He2016ResNet}. We use Adam (learning rate $10^{-3}$, weight decay $5\times10^{-4}$) and batch size 128. Unless otherwise specified, bias is applied only to the target class \texttt{automobile} (class index 1).

\textbf{Two-stage protocol.} The Clean stage trains on clean data for 30 epochs to establish a baseline. During Bias Exposure, the specified bias is continuously introduced for 10, 20, 30, or 50 epochs. Each configuration uses five random seeds (0--4).

\textbf{Bias mechanisms.} We cover static and dynamic mechanisms. Structural Patch represents an acquisition-related shortcut, such as one arising from differences in imaging equipment, hospitals, or acquisition workflows. A model may then rely on a proxy feature that is unrelated to class semantics but correlated with the label.

Static bias has three forms. Structural Patch applies a fixed $4\times4$ patch of magnitude $+0.10$ to the lower-right corner of automobile images (enabled only during training). Label Flip randomly flips the labels of automobile examples with probability $p=0.5$. Feature Corruption adds Gaussian noise ($\sigma=0.3$) to automobile inputs.

Dynamic bias has two forms. Prediction-Dependent Label Flip applies prediction-dependent flipping with $p=0.5$ only to automobile examples: if the current prediction is correct, it is flipped to the highest-probability incorrect class; if it is wrong, it is flipped to the current predicted class to reinforce the error. This creates a $\text{Model} \rightarrow \text{Label} \rightarrow \text{Model}$ feedback loop. Self-Induced Data Drift performs inference on the full training set after each Bias epoch, selects the $7.5\%$ of examples with the smallest margin (the difference between the top-1 and top-2 probabilities), and changes their labels to the model's top-1 predictions. Changes accumulate during Bias Exposure, creating a stronger $\text{Model} \rightarrow \text{Data} \rightarrow \text{Model}$ loop.

\textbf{Evaluation metrics.} We use test accuracy on the automobile subset as the primary correctness measure. Let $R_{\mathrm{clean}}$ be the mean automobile accuracy over the final five Clean epochs (25--29), and let $R_{\mathrm{bias-end}}$ be accuracy at the end of Bias Exposure. For Self-Induced Drift, we additionally report the cumulative modification fraction and test-set feature effective rank (eRank) at the end of Bias Exposure.

\subsection*{CIFAR-10 results}

\subsubsection*{Static bias}

Table~\ref{tab:cifar_static} reports target-class accuracy at the end of Bias Exposure under three static biases. With Structural Patch, mean $R_{\mathrm{bias-end}}$ falls from $78.12\pm10.73\%$ at Exposure $=10$ to the range $57.80\%$--$61.92\%$ at Exposure $=20$, $30$, and $50$. Compared with the same-seed Clean reference, a decrease of about 28--32 percentage points is already present at Exposure $=20$. Bias-end accuracy under Label Flip and Feature Corruption remains mostly in the $90\%$--$93\%$ range. The available summary files do not retain Clean references for these two mechanisms, so paired decreases are not calculated in the table.

\begin{table}[ht]
\centering
\small
\setlength{\tabcolsep}{3pt}
\caption{Target-class accuracy at the end of Bias Exposure under static bias on CIFAR-10 (five seeds, mean $\pm$ standard deviation). The decrease from Clean is computed as $R_{\mathrm{clean}}-R_{\mathrm{bias-end}}$ for the same seed; “--” indicates that the corresponding Clean reference was not saved.}
\label{tab:cifar_static}
\begin{tabularx}{\linewidth}{l c c X}
\hline
Mechanism & Exposure & \makecell{Bias-end accuracy (\%)} & \makecell{Decrease from Clean\\(percentage points)} \\
\hline
Structural Patch & 10 & $78.12\pm10.73$ & $11.51\pm10.73$ \\
Structural Patch & 20 & $61.68\pm18.49$ & $27.95\pm19.42$ \\
Structural Patch & 30 & $57.80\pm16.69$ & $31.83\pm17.04$ \\
Structural Patch & 50 & $61.92\pm8.13$ & $28.64\pm7.74$ \\
\hline
Label Flip & 10 & $91.92\pm3.02$ & -- \\
Label Flip & 20 & $93.06\pm3.84$ & -- \\
Label Flip & 30 & $93.60\pm3.10$ & -- \\
Label Flip & 50 & $92.68\pm1.40$ & -- \\
\hline
Feature Corruption & 10 & $91.44\pm3.34$ & -- \\
Feature Corruption & 20 & $89.88\pm3.86$ & -- \\
Feature Corruption & 30 & $93.32\pm2.61$ & -- \\
Feature Corruption & 50 & $93.08\pm2.00$ & -- \\
\hline
\end{tabularx}
\end{table}

\subsubsection*{Dynamic bias and self-induced drift}

Table~\ref{tab:cifar_dynamic} reports target accuracy at the end of Prediction-Dependent Label Flip. Across exposure durations, mean Bias-end accuracy over five seeds ranges from $22.98\%$ to $49.62\%$, with substantial variation across seeds and no monotonic decrease as Exposure length increases.

\begin{table}[ht]
\centering
\caption{Target accuracy at the end of Bias Exposure under Prediction-Dependent Label Flip on CIFAR-10 (five seeds, mean $\pm$ standard deviation).}
\label{tab:cifar_dynamic}
\begin{tabular}{c|c}
\hline
Exposure duration & Bias-end accuracy (\%) \\
\hline
10 & $49.62\pm32.78$ \\
20 & $24.44\pm19.41$ \\
30 & $39.76\pm31.13$ \\
50 & $22.98\pm26.89$ \\
\hline
\end{tabular}
\end{table}

Table~\ref{tab:cifar_self_induced} reports the state at the end of Bias Exposure under Self-Induced Data Drift. The cumulative modification fraction increases from about $36\%$ to $70\%$ as Exposure increases, while Bias-end eRank falls from 36.2 to 29.3. Meanwhile, target accuracy remains above $90\%$ in most conditions. These results document closed-loop data drift and representational change; we do not interpret them as showing continued deterioration in target accuracy.

\begin{table}[ht]
\centering
\small
\setlength{\tabcolsep}{3pt}
\caption{Closed-loop effects of Self-Induced Data Drift on CIFAR-10 at the end of Bias Exposure (five seeds; accuracy is mean $\pm$ standard deviation).}
\label{tab:cifar_self_induced}
\begin{tabularx}{\linewidth}{c X c c}
\hline
Exposure & \makecell{Cumulative modification\\fraction} & Bias-end eRank & \makecell{Bias-end accuracy (\%)} \\
\hline
10 & $0.361\pm0.006$ & $36.2$ & $93.68\pm2.28$ \\
20 & $0.499\pm0.003$ & $34.0$ & $90.82\pm2.11$ \\
30 & $0.584\pm0.008$ & $31.6$ & $92.66\pm1.80$ \\
50 & $0.698\pm0.010$ & $29.3$ & $91.88\pm1.38$ \\
\hline
\end{tabularx}
\end{table}

\subsection*{Cross-dataset replication on Tiny-ImageNet}

To test whether the SBW pattern under persistent bias replicates across datasets, we evaluate Structural Patch and Self-Induced Drift on Tiny-ImageNet-200 ($64\times64$, 200 classes). The model is a ResNet-18 with AdaptiveAvgPool and a 200-way output \cite{He2016ResNet}, trained with Adam. The Clean stage lasts 30 epochs; Bias Exposure lasts 10, 20, 30, or 50 epochs, with five seeds per condition. The target class is sports car (WordNet \texttt{n04285008}), corresponding to automobile in CIFAR-10. Structural Patch uses a lower-right $+0.10$ patch with area matched by relative image size (size $8\times8$). Self-Induced Drift changes the $7.5\%$ of examples with the smallest margin after each Bias epoch and accumulates the changes. Target-class accuracy is based on 50 validation images in that class, so individual results are variable.

Table~\ref{tab:tiny_bias_end} summarizes the Bias-end states of both mechanisms. The five-seed mean target accuracy under Structural Patch ranges from $34.40\%$ to $51.20\%$; means at 30 and 50 epochs are below those at 10 and 20 epochs, although variation across seeds remains. Under Self-Induced Drift, the cumulative modification fraction rises from about $43\%$ to $81\%$, while Bias-end eRank falls from $162.5$ to $116.9$. These results assess replication across datasets.

\begin{table}[ht]
\centering
\caption{Results at the end of Bias Exposure under different persistent-bias mechanisms on Tiny-ImageNet-200 (five seeds, mean $\pm$ standard deviation). eRank and cumulative modification fraction are reported only for Self-Induced Drift.}
\label{tab:tiny_bias_end}
{\small
\setlength{\tabcolsep}{3pt}
\begin{tabularx}{\linewidth}{X c c c c}
\hline
Mechanism & Exposure & \makecell{Cumulative\\modification (\%)} & \makecell{Bias-end\\eRank} & \makecell{Bias-end accuracy\\(\%)} \\
\hline
Structural Patch & 10 & -- & -- & $46.40\pm14.10$ \\
Structural Patch & 20 & -- & -- & $51.20\pm10.83$ \\
Structural Patch & 30 & -- & -- & $34.40\pm12.60$ \\
Structural Patch & 50 & -- & -- & $35.60\pm5.37$ \\
\hline
Self-Induced Drift & 10 & $43.32\pm0.32$ & $162.5\pm8.3$ & $44.40\pm23.64$ \\
Self-Induced Drift & 20 & $60.68\pm0.40$ & $144.4\pm9.2$ & $58.80\pm7.69$ \\
Self-Induced Drift & 30 & $70.30\pm0.47$ & $132.0\pm8.7$ & $44.40\pm16.64$ \\
Self-Induced Drift & 50 & $81.11\pm0.52$ & $116.9\pm5.7$ & $41.60\pm23.60$ \\
\hline
\end{tabularx}
}
\end{table}

\subsection*{Summary}

The CIFAR-10 and Tiny-ImageNet results show that persistent bias can change target performance or training-data state at the end of Bias Exposure, but the change depends on the bias mechanism and task conditions. The Tiny-ImageNet experiment tests whether this SBW pattern replicates across datasets.

\clearpage
\section{Challenge--Correction Experimental Details}
\label{app:challenge-correction}

This appendix reports the environment construction, run selection, interventions, and evaluation metrics for the two validation experiments in Sec.~\ref{sec:challenge-correction}. Both assume that external ground truth and the judgment to be tested are specified; they assess whether a known error can be confirmed and corrected.

\subsection{Controlled scientific-discovery experiment}

\subsubsection{Environment and candidate regions}

The experiment is defined in the five-dimensional search space $[0,1]^5$. Each location represents a candidate solution, and its true utility is given by a prespecified deterministic function with values in $[0,1]$. The true utility function has three main high-value regions: region A has a peak near 0.7, is relatively broad, and is easy for autonomous search to discover; region B has a peak of 1.0 but is narrower and more remote; and region C has a peak near 0.9, also above A. Both model judgments and external evaluation use the mean over prespecified regional neighborhoods rather than the single value at the peak. In the Stable-but-Wrong environment, the true mean is approximately 0.67 in the neighborhood of A and 0.73 in the neighborhood of B, so the true ranking is $B>A$.

\subsubsection{Historical records and run selection}

Each run first observes 32 initial locations, then conducts 30 rounds of autonomous search, adding 10 locations per round. The final history for each run therefore contains
\begin{equation}
    32 + 30\times10 = 332
\end{equation}
observations. Response values in the history are retained as the noisy observations actually obtained during learning; they are not recomputed or replaced using the true utility function.

The initial experiment contains 20 runs numbered 0--19. A run is included in the Challenge--Correction analysis only if it satisfies both conditions: first, at the end of search the model still ranks the A neighborhood above the B neighborhood; second, the search has not discovered another region with true utility higher than A. Run 19 discovered region C and was excluded, leaving 19 runs in the final analysis.

\subsubsection{Challenge and matched control}

The A and B evaluation locations are fixed before analysis, rather than selected post hoc from each run's 332 historical observations. For each run, we compute the model's mean predictions $\widehat{\mu}_A$ and $\widehat{\mu}_B$ and the corresponding true regional means $\mu_A^*$ and $\mu_B^*$. We classify the judgment as SBW when
\begin{equation}
    \widehat{\mu}_A > \widehat{\mu}_B,
    \qquad
    \mu_A^* < \mu_B^*.
\end{equation}

The matched Normal environment shares the same 332 historical locations and observations as the Stable-but-Wrong environment, but its external true structure is defined so that A is indeed better than B. This control tests whether Challenge misclassifies a correct conclusion and whether a subsequent intervention incorrectly flips a correct ranking.

In all 19 included runs, the model's ranking disagrees with the true ranking in the SBW environment and agrees with it in the Normal environment. Because inclusion requires the model to prefer A at the end of search, the 19/19 result checks a prespecified judgment against ground truth; it is not an estimate of automatic-detection accuracy across unknown errors.

\subsubsection{Correction protocol and endpoint}

Each intervention starts independently from the same frozen 332-point history; data are not accumulated from one intervention to another. The number of added observations is $k\in\{1,2,5,10\}$, at most about 3\% of the original history. All added locations receive noise-free scores from the true utility function; locations are frozen before scores are read. The intervention locations are selected in three ways: randomly from the search space, from the B neighborhood, or under an oracle-center upper-bound condition, with the experimenter-known center of B as the first point and the remaining points in a prespecified tiny neighborhood. The model is refit after new observations are added, while the original 332 noisy observations remain unchanged.

The primary endpoint is the regional prediction difference after refitting:
\begin{equation}
    m_{BA}=\widehat{\mu}_B-\widehat{\mu}_A.
\end{equation}
$m_{BA}\leq0$ means that the model still prefers A, whereas $m_{BA}>0$ means that it now prefers B. In the Stable-but-Wrong environment, a change from $m_{BA}\leq0$ to $m_{BA}>0$ is counted as a correction; the same change in the Normal environment is counted as an erroneous flip.

\begin{table}[H]
    \centering
    \caption{Conclusion flips in the controlled scientific-discovery experiment. The denominator is 19 paired runs.}
    \label{tab:synthetic-correction}
    \begin{tabular}{lcccc}
        \hline
        Intervention arm / environment & $k=1$ & $k=2$ & $k=5$ & $k=10$ \\
        \hline
        Random true observations / SBW       & 0/19  & 0/19  & 0/19  & 0/19  \\
        True observations from B neighborhood / SBW & 13/19 & 16/19 & 19/19 & 19/19 \\
        B-center upper bound / SBW          & 19/19 & 19/19 & 19/19 & 19/19 \\
        Random true observations / Normal    & 0/19  & 0/19  & 0/19  & 0/19  \\
        True observations from B neighborhood / Normal   & 0/19  & 0/19  & 0/19  & 0/19  \\
        B-center upper bound / Normal       & 0/19  & 0/19  & 0/19  & 0/19  \\
        \hline
    \end{tabular}
\end{table}

The overall correction rate for B-neighborhood observations rises with $k$, although individual runs need not change monotonically. In one run, the model switches to B at $k=1$, back to A at $k=2$, and to B again at $k=5$. Larger budgets therefore improve correction reliability overall, but not in every trajectory. Relevant observations improve the local A--B margin in all 19 runs, while global prediction error on held-out locations does not improve in parallel. The result concerns the specified regional ranking and does not imply improved overall model accuracy. Paired McNemar tests of flip events between the SBW and Normal worlds give $p=0.000244$, $0.0000305$, $0.00000381$, and $0.00000381$ for B-neighborhood observations at $k=1,2,5,10$, respectively; for the B-center upper-bound condition, $p=0.00000381$ at every budget.

\subsection{CIFAR-10 Structural Patch experiment}

\subsubsection{Construction of erroneous dependence}

The experiment uses CIFAR-10 and a ResNet-18 classifier. During Bias Exposure in the Structural Patch condition, we add a fixed $4\times4$ marker of magnitude $+0.10$ to the lower-right corner of automobile images. Image labels remain unchanged. Because the marker is added by the experimenter and contains no class semantics, its association with the automobile label is known to be erroneous.

The formal Challenge--Correction experiment uses five independent random seeds. The Wrong condition starts from the checkpoint at the end of Bias Exposure with patch magnitude $+0.10$; the paired Correct condition starts from the corresponding clean checkpoint before Bias Exposure. Magnitude $+0.10$ was selected using a prespecified rule: the mean increase in shortcut sensitivity on preview seeds must be at least 0.30, while mean automobile-class accuracy after Bias Exposure must be at least 40\%. Magnitudes $+0.05$ and $+0.15$ are used only as descriptive strength checks.

\subsubsection{Challenge metric}

Shortcut sensitivity is computed on a frozen set of 9000 non-automobile test images. For each image, we obtain the prediction without the marker and with the same patch added, and define
\begin{equation}
    S_{\mathrm{shortcut}}
    =
    \Pr\!\left[f(x,\mathrm{patch})\neq f(x,\mathrm{clean})\right].
\end{equation}
This metric is the proportion of predictions that change when only the known non-semantic marker is changed. Mean $S_{\mathrm{shortcut}}$ is 0.335 for the Wrong checkpoint and 0.012 for the paired clean checkpoint; mean automobile-class accuracy for the Wrong checkpoint is 48.3\%. In preview experiments, the mean sensitivity increases relative to the clean checkpoint by 0.127, 0.327, and 0.418 at magnitudes $+0.05$, $+0.10$, and $+0.15$, respectively.

\subsubsection{Correction protocol}

The update budgets are $k\in\{10,50,200\}$. We compare four conditions: no update, random-sample control, matched cue-absent control, and relevant counterexamples. In the relevant arm, about half of the evidence consists of non-automobile images with the same patch added while retaining the true labels; the other half consists of clean automobile images. The matched cue-absent control uses exactly the same image indices, class composition, and true labels, but does not add the patch to non-automobile images. The random-sample control uses stratified samples with the same budget. This design compares evidence that breaks the association “patch implies automobile” with ordinary updates.

Each update consists of five epochs, with 5000 stratified replay samples per epoch. We use Adam with learning rate $3\times10^{-4}$ and batch size 128, and freeze batch-normalization statistics. We also add a logit-distillation constraint with temperature 2 and weight $\lambda=0.1$. Each update condition includes five source seeds and three fixed repeats. Challenge test images are not used for updates.

The primary prespecified comparison is the difference in change in $S_{\mathrm{shortcut}}$ between relevant counterexamples and the matched cue-absent control at $k=50$. A negative value indicates reduced dependence on the patch.

\begin{table}[H]
    \centering
    \caption{Change in shortcut sensitivity in the CIFAR-10 Wrong condition. Paired advantage is the additional reduction from relevant counterexamples relative to the matched cue-absent control.}
    \label{tab:cifar-correction}
    \begin{tabular}{rcccc}
        \hline
        $k$ & Relevant counterexamples & Matched cue-absent & Random control & Paired advantage \\
        \hline
        10  & $-0.137$ & $-0.088$ & $-0.084$ & 0.050 \\
        50  & $-0.234$ & $-0.087$ & $-0.092$ & 0.146 \\
        200 & $-0.277$ & $-0.078$ & $-0.081$ & 0.199 \\
        \hline
    \end{tabular}
\end{table}

At the prespecified $k=50$ condition, relevant counterexamples improve over the matched control in the same direction for all five source seeds, with a mean paired advantage of 0.146 (one-sided exact Wilcoxon $p=0.03125$; seed-level bootstrap 95\% CI $[0.120,0.177]$). Relevant counterexamples reduce $S_{\mathrm{shortcut}}$ from 0.335 to about 0.101 and raise mean automobile-class accuracy from 48.3\% to 80.5\%. The intervention therefore substantially weakens, but does not eliminate, erroneous dependence; post-update sensitivity remains above the clean checkpoint value of 0.012.

\subsubsection{Safety check in the Correct condition}

The same updates are also applied to the paired Correct checkpoint. At none of the prespecified budgets does any run show an overall test-accuracy decrease greater than one percentage point. Under the relevant-counterexample condition, the mean false-correction rate across 45 updates is 0.0142, the worst single-class accuracy change is $-0.74$ percentage points, and mean shortcut sensitivity after updates remains near 0.010. The paired comparison uses a diagnostic suite that had been used to select hyperparameters in an earlier protocol, so its results should be interpreted as within-experiment paired differences; the formal Wrong checkpoint at magnitude $+0.10$ was trained under an independent magnitude-selection protocol. These results show only that we observe no evident harm under the current dataset, model, patch, and update protocol; they do not provide a general safety guarantee.

\subsection{Scope of interpretation}

Both experiments are conditional minimal validations. The controlled scientific-discovery experiment assumes that the researcher already knows to compare A and B and can query the true utility function. The CIFAR-10 experiment assumes that the researcher already knows the Structural Patch is an erroneous feature unrelated to class and can construct counterexamples that break its label association. The experiments therefore support the claim that stable errors can be confirmed and corrected or reduced by targeted evidence when external ground truth and the error to be tested are known. They do not test automatic error discovery without ground truth, nor do they support the claim that arbitrary additional data or interventions would have the same effect.

\section{Theoretical Foundation of Stable but Wrong}
\label{app:theory}

We use a minimal optimization model to show how persistent bias can separate stable convergence from correctness.

Consider the following learning update:
\[
\theta_{t+1}
=
\theta_t
-
\eta
\left(
\nabla L(\theta_t)+b_t
\right),
\]
where $L(\theta)$ is the original learning objective aligned with the true objective, $\nabla L(\theta_t)$ is its optimization direction, $\eta>0$ is the learning rate, and $b_t$ is a systematic bias that persists in the learning process.

The bias term $b_t$ represents persistent bias from sources such as rewards, gradient estimates, labels, or feedback. We do not model the full non-convex dynamics of a particular deep-learning system. Instead, we ask whether persistent bias can separate stable convergence from the true objective. Gradient descent may operate normally and converge to the signal it receives; when that signal is biased, convergence does not guarantee correctness relative to the true objective \cite{Bottou1998Online,Bottou2018Optimization}.

\paragraph{Proposition 1 (Existence of Stable but Wrong).}

Let $L(\theta)$ be a continuously differentiable, strongly convex function with a unique minimizer
\[
\theta^*
=
\arg\min_{\theta}L(\theta).
\]
Assume that $\nabla L(\theta)$ is Lipschitz continuous, and choose a constant learning rate $\eta>0$ such that the gradient-descent map $G(\theta)=\theta-\eta\nabla L(\theta)$ is a contraction \cite{Nesterov2004Convex}. If the persistent bias satisfies
\[
b_t\rightarrow b,\qquad b\neq0,
\]
then the biased iteration converges to a unique stable point $\theta_{\infty}$ satisfying
\[
\nabla L(\theta_{\infty})+b=0,
\]
and
\[
\theta_{\infty}\neq\theta^*.
\]

\paragraph{Proof sketch.}

Define the shifted objective
\[
\widetilde L_b(\theta)=L(\theta)+b^\top\theta.
\]
Adding a linear term preserves strong convexity and leaves gradient differences unchanged. Thus, the associated map
\[
G_b(\theta)=\theta-\eta\left(\nabla L(\theta)+b\right)
\]
has the same contraction modulus as $G$ and has a unique fixed point $\theta_b$ satisfying
\[
\nabla L(\theta_b)+b=0.
\]
Let $e_t=b_t-b$. The original iteration can be written as
\[
\theta_{t+1}=G_b(\theta_t)-\eta e_t,
\qquad e_t\rightarrow0.
\]
Because the additive perturbation vanishes asymptotically, the contractive iteration converges to $\theta_b$. At the true minimizer, $\nabla L(\theta^*)=0$, whereas at $\theta_b$, $\nabla L(\theta_b)=-b\neq0$. Thus $\theta_b\neq\theta^*$ and $\theta_{\infty}=\theta_b$: the dynamics converge normally to a point that is wrong relative to the original objective.

\paragraph{Corollary 1.}

In the presence of persistent bias in learning, internal stability measures alone---such as parameter changes, gradient norms, stable training loss, or policy-update magnitudes---cannot guarantee that a learning system is close to the true objective \cite{ShalevShwartz2010Learnability}.

\paragraph{Implication.}

Proposition 1 does not describe the full dynamics of modern deep-learning systems, whose loss landscapes are typically non-convex and whose representations evolve in complex ways. Strong convexity provides a minimal existence result: even in this simple setting, persistent bias can separate the converged state from the true objective. Proposition 1 establishes only the existence of an SBW state.

\end{appendices}
\bibliographystyle{unsrt}


\end{document}